# Leveraging Large Language Models for Rare Disease Named Entity Recognition

Nan Miles Xi [1], Yu Deng [1], Lin Wang [2*]

[1] Data and Statistical Sciences, AbbVie Inc., North Chicago, IL 60064, USA

[2] Department of Statistics, Purdue University, West Lafayette, IN 47907, USA

* Correspondence: linwang@purdue.edu

## Abstract

Named Entity Recognition (NER) in the rare disease domain poses unique challenges due to limited labeled data, semantic ambiguity between entity types, and long-tail distributions. In this study, we evaluate the capabilities of GPT-4o for rare disease NER under low-resource settings, using a range of prompt-based strategies including zero-shot prompting, few-shot in-context learning, retrieval-augmented generation (RAG), and task-level fine-tuning. We design a structured prompting framework that encodes domain-specific knowledge and disambiguation rules for four entity types. We further introduce two semantically guided few-shot example selection methods to improve in-context performance while reducing labeling effort. Experiments on the RareDis Corpus show that GPT-4o achieves competitive or superior performance compared to BioClinicalBERT, with task-level fine-tuning yielding the strongest performance among the evaluated approaches and improving upon the previously reported BioClinicalBERT baseline. Cost-performance analysis reveals that few-shot prompting delivers high returns at low token budgets. RAG provides limited overall gains but can improve recall for challenging entity types, especially signs and symptoms. An error taxonomy highlights common failure modes such as boundary drift and type confusion, suggesting opportunities for post-processing and hybrid refinement. Our results demonstrate that prompt-optimized LLMs can serve as effective, scalable alternatives to traditional supervised models in biomedical NER, particularly in rare disease applications where annotated data is scarce.

## 1 Introduction

Rare diseases are individually rare but collectively common, with over 6,000 distinct conditions affecting an estimated 300 million people worldwide [1]. Their low prevalence means that general practitioners have little experience with any given rare disease, while the clinical heterogeneity across conditions further complicates diagnosis [2]. As a result, patients often face prolonged processes before receiving a correct diagnosis and appropriate treatment [3]. This diagnostic gap has elevated rare diseases to a global health priority and highlights the urgent need for scalable methods to extract and disseminate rare disease knowledge. Automated information extraction, particularly named entity recognition (NER), can play a pivotal role in addressing this gap. NER enables the construction of biomedical knowledge graphs linking diseases to phenotypes, supports clinical decision-making, and assists patient care by surfacing relevant findings in medical narratives [4–6]. Machine learning decision support models have been explored to improve diagnostic accuracy using clinical data [7], but such approaches often depend on reliable extraction of disease and phenotype information from unstructured notes, motivating robust clinical NER. Recent work has also demonstrated the utility of NER for symptom surveillance in social media platforms [8]. However, extracting such information from unstructured text poses several challenges.

Foremost among these is the low-resource setting. Few annotated corpora exist for rare disease NER, as expert labeling is costly and time-consuming. In addition, rare disease terminology is often semantically ambiguous, which creates overlapping entity boundaries and introduces high annotation variability. Distinguishing between such entities requires nuanced domain understanding that even advanced models struggle to achieve [9]. Compounding these issues is the long-tail distribution of rare diseases. The vast majority occur with low frequency, often below one case per million individuals [10]. Consequently, most rare disease mentions appear infrequently in existing data, and language models may lack sufficient exposure to ultra-rare conditions. Any robust solution must contend with both data scarcity and domain-specific ambiguity to succeed in this setting.

Conventional biomedical NER systems rely on supervised learning with domain-specific models. Transformer-based architectures such as BioClinicalBERT and BioBERT have achieved strong performance on medical NER tasks when trained on large-scale corpora [11,12]. However, these supervised approaches are inherently constrained by their dependence on large and high-quality annotated datasets, which remain scarce in the rare disease domain. Even when such data are available, generalization to novel or ultra-rare entities remains difficult due to the long-tail distribution of biomedical concepts [13]. In parallel, generative large language models (LLMs) have enabled a shift toward prompt-based learning through natural language instructions. In general-domain applications, generative LLMs have demonstrated impressive zero-shot and few-shot capabilities, substantially reducing the need for task-specific labeled data [14]. Several recent studies have begun to test prompt-engineering for rare-disease extraction [3,15–17], yet systematic evaluation and broader generalization remain open questions.

Prompt-based NER in biomedical text introduces several open questions. Early evaluations indicate that general-purpose LLMs underperform compared to fine-tuned biomedical models on token-level clinical NER tasks [18,19]. Moreover, prompt designs tailored to rare disease extraction are still in their infancy. It remains unclear whether off-the-shelf LLMs can reliably disambiguate the subtle semantic distinctions in rare disease contexts. Beyond basic prompting, two complementary approaches, retrieval-augmented generation (RAG) and in-context learning, offer potential solutions. RAG enables an LLM to access external information at inference time by retrieving and incorporating supporting documents [20]. In rare disease NER, RAG can allow the model to consult definitions or explanations from curated biomedical knowledge databases. Similarly, the effectiveness of in-context learning relies on the choice of labeled learning exemplars. Recent studies have shown that selecting semantically similar examples can substantially improve few-shot learning in biomedical Natural Language Processing (NLP) tasks [21]. Yet, it remains unclear how example selection strategies affect LLM performance in disambiguating complex rare disease entities.

Given these challenges, we aim to answer the following question: Can generative LLMs accurately and cost-effectively perform NER in the rare disease domain using prompt-based methods, fine-tuning on domain-specific prompts, or retrieval-augmented context? We focus on OpenAI's GPT-4o as a representative LLM [22] and evaluate its ability to identify rare disease-related entities under various low-resource settings. We benchmark GPT-4o against BioClinicalBERT to quantify the strengths and limitations of prompt-based LLMs in specialized biomedical tasks. BioClinicalBERT is a strong domain specific transformer pretrained on biomedical literature and clinical notes, widely used as a state-of-the-art (SOTA) baseline for clinical and biomedical NLP. Our goal is to assess whether prompt-only, in-context learning, and RAG can approach SOTA models without large and annotated datasets, and to understand their trade-offs relative to traditional supervised learning approaches.

Our contributions in this paper are summarized as follows. First, we design a prompt template that encodes domain knowledge for semantically overlapping entity types. This framework guides GPT-4o to perform entity recognition with nuanced semantic boundaries. Second, we evaluate GPT-4o under multiple prompting regimes and compare its performance against the SOTA. This comparison quantifies the effectiveness of prompt-based LLMs relative to conventional NER systems. Third, we investigate two context-aware strategies for selecting learning examples. We show that these methods outperform random selection and enhance GPT-4o's ability to resolve ambiguous entity mentions. Another contribution is that we implement an RAG approach and let GPT-4o retrieve contextual snippets from a knowledge base. We assess the utility of this external biomedical context and highlight when RAG provides meaningful performance gains. We also evaluate the inference cost across different prompting strategies. This analysis provides insights into the deployment feasibility of prompt-based LLMs in real-world applications. Finally, we introduce a taxonomy of error types and perform an error analysis to identify common failures in GPT-4o's output.

In our experiments, GPT-4o demonstrates strong performance on rare disease NER under minimal supervision. With a small number of in-context learning examples, GPT-4o's performance approaches that of the fine-tuned BioClinicalBERT. Importantly, we find that the quality of selected examples plays a critical role in this success. Semantic selection strategies consistently outperform random selection by enabling the model to resolve ambiguous entity boundaries and improve recall. In contrast, RAG provides only marginal benefits. Overall, our findings indicate that prompt-engineered LLMs can deliver competitive NER performance in the rare disease domain. However, challenges remain in disambiguating closely related entity types and addressing edge cases with low frequency. Our error analysis reveals systematic failure modes, with most errors stemming from span boundary mismatches. These insights highlight specific areas for future refinement, such as post-processing heuristics and hybrid LLM rule-based systems to improve boundary resolution and type specificity [23,24].

The remainder of this paper is structured as follows. Section 2 describes the methodology, including the rare disease dataset, prompt design, learning example selection, and RAG components. Section 3 presents the experimental results and evaluation, performance comparisons, ablation studies, and error analysis. Section 4 discusses the implications of these findings and concludes the paper with future directions.

# 2 Methods

## 2.1 RareDis Corpus Dataset

We utilize the RareDis Corpus, a domain-specific dataset developed to support NLP applications in the rare disease domain [25]. Let the dataset be denoted as:

$$\mathbf{D} = \{(x_i, Y_i)\}_{i=1}^{N}$$

where $x_i \in \mathbf{X}$ is a biomedical document and $Y_i = \{(s_{ij}, c_{ij})\}_{j=1}^{n_i} \subset \mathbf{Y}$ is the set of annotated entities, with $s_{ij}$ denoting a surface text span, $n_i$ being the number of annotated entities in document $x_i$, and $\tau_{ij} \in \boldsymbol{\tau}$ representing the entity type. The entity space is defined as:

$$\boldsymbol{\tau} = \{\text{rare disease}, \text{disease}, \text{sign}, \text{symptom}\}$$

The corpus contains $N = 1{,}041$ documents sourced from the National Organization for Rare Disorders (NORD) database [26]. Each document is structured into multiple clinically relevant sections, including general discussion, signs and symptoms, causes, diagnosis, related disorders, affected populations, and therapies. Entity annotations are performed manually by domain experts and contain 5,221 rare disease mentions, 2,348 general disease mentions, 5,333 signs, and 396 symptoms. The corpus is split into training (70%), validation (10%), and test (20%) subsets:

$$\mathbf{D} = \mathbf{D}_{\text{train}} \cup \mathbf{D}_{\text{val}} \cup \mathbf{D}_{\text{test}}$$

As shown in **Table 1**, the entity distribution is highly imbalanced across types. For example, symptoms are the sparsest category in the test set (n = 53). This imbalance, together with the long-tail nature of rare disease concepts, motivates imbalance-aware training and data augmentation strategies as important directions for future work.

The RareDis Corpus reports an Inter-Annotator Agreement (IAA) with an average F1 score of 83.5% for entity recognition, reflecting a high degree of annotation consistency [25]. The corpus makes fine-grained distinctions between semantically related entity types: disease vs. rare disease (based on prevalence thresholds) and sign vs. symptom, where signs are objective clinician-observable findings (e.g., physical exam, labs, imaging) and symptoms are subjective patient-reported experiences (e.g., pain, fatigue, nausea). These subtle boundaries introduce substantial challenges for LLMs in entity recognition. A detailed breakdown of the entity statistics, definitions, and representative examples is provided in **Table 1**.

## 2.2 Model and Prompt Design

We utilize OpenAI's pretrained large language model GPT-4o to perform NER in the rare disease domain, treating the task as conditional sequence generation [27]. For each test input $x_{\text{inquiry}} \in \mathbf{X}_{\text{test}}$, the model is provided with a prompt $\pi \in \mathbf{P}$, constructed from five structured components designed to instruct the model on entity recognition without labeled training examples:

$$\pi = \text{task description} \parallel \text{output format} \parallel \text{task guidance} \parallel \text{disambiguation rule} \parallel x_{\text{inquiry}}$$

Here, the components are defined as follows:

- **Task description** specifies the recognition objective, denoted by a label $\tau \in \boldsymbol{\tau}$. For instance: "*Identify the names of rare diseases from the following text*".
- **Output format** enforces a standardized, comma-separated list of identified entities $\hat{y}_{\text{inquiry}} \in \mathbf{Y}$, enabling exact-match evaluation. For example: "*Output only the exact disease names without any additional changes. If there are multiple diseases, separate their names with commas. If there is no disease, output none*".
- **Task guidance** provides formal definitions for each entity type $\tau$, assisting the model to distinguish between semantically overlapping categories. For example: "*Symptoms are subjective experiences reported by the patient, which cannot be directly observed or measured by others. They reflect what the patient feels, such as pain, fatigue, or nausea. Symptoms are experienced internally and rely on the patient's description*".
- **Disambiguation rule** offers meta-instructions highlighting frequent errors observed during validation. These discourage undesirable behaviors such as misclassifying general diseases as rare diseases or merging distinct entities. For example: "*Treat abbreviations as separate rare disease names. Do not identify regular diseases as rare diseases.*"

- **Input text** ($x_{\text{inquiry}}$) is the raw contents from which entities are to be identified. A prefix marks its beginning, such as: "*The text from which you need to extract the signs of rare diseases is: …*"

We define the basic prompt components as the combination of the task description, output format, and $x_{\text{inquiry}}$. Basic prompt contains the core instruction and context. Advanced components include task guidance and disambiguation rules, which encode domain knowledge and observed failure modes. All prompts are constructed without including any labeled examples (i.e., zero-shot learning), ensuring that the model's performance is attributable solely to prompt content and pretrained knowledge. To quantify the contribution of each prompt category, we vary the presence of basic and advanced components in the complete prompt π and evaluate the zero-shot performance of GPT-4o under each configuration using the evaluation framework described in Section 2.6. A complete set of prompt templates by entity type is summarized in **Table 2**.

### 2.3 In-Context Learning and Example Selection Strategies

In-context learning refers to providing demonstration examples directly in the prompt to guide the model's response, without gradient-based parameter updates [14,28]. Formally, let $\{(x_i, y_i)\}_{i=1}^{k} \subset \mathbf{D}_{\text{train}}$ denote a set of $k$ in-context learning examples, where $x_i \in \mathbf{X}_{\text{train}}$ and $y_i \in \mathbf{Y}_{\text{train}}$. Here, $y_i$ is a flattened, comma-separated list of entities derived from the structured annotations $Y_i$. LLM receives a prompt of the form:

$$\pi = \text{basic components} \parallel \text{advanced components} \parallel \left\{\left(x_j, y_j\right)\right\}_{j=1}^{k} \parallel x_{\text{inquiry}}$$

The model then generates output $\hat{y}_{\text{inquiry}} = M(\pi)$, where $M$ is the LLM conditioned on the full prompt. Depending on $k$, the setup is referred to as one-shot ($k = 1$) or few-shot ($k > 1$) learning. To assess how different configurations of in-context learning examples affect model performance, we explore a set of example selection methods by leveraging semantic similarity between input texts. Each $x_{\text{inquiry}} \in \mathbf{X}_{\text{test}}$ is mapped to an embedding vector $f(x) \in \mathbf{R}^{3072}$ using OpenAI's text-embedding-3-large model. Given two texts $x$ and $x'$, semantic similarity is quantified via the Euclidean distance:

$$d(x, x') = \|f(x) - f(x')\|_2$$

We then consider the following three selection strategies:

- **Inquiry-Random** – For each $x_{\text{inquiry}}$, select $k$ learning examples uniformly at random from $\mathbf{X}_{\text{train}}$ independent of semantic similarity.

- **Inquiry-KNN** – For each $x_{\text{inquiry}}$, compute $d(x_{\text{inquiry}}, x_i)$ for all $x_i \in \mathbf{X}_{\text{train}}$, and select the top $k$ learning examples with the smallest distances. This yields context-specific, nearest-neighbor demonstrations.
- **Cluster-KNN** – Partition the test set $\mathbf{X}_{\text{test}}$ into $C$ clusters using k-means clustering in the embedding space. Let $\mathbf{C}_j \subset \mathbf{X}_{\text{test}}$ denote the set of inquiry texts in cluster $j$. For each training example $x_i \in \mathbf{X}_{\text{train}}$, define its average distance to cluster $j$ as:

$$\bar{d}_j(x_i) = \frac{1}{|\mathbf{C}_j|} \sum_{x \in \mathbf{C}_j} d(x, x_i)$$

  Then, for every $x_{\text{inquiry}} \in \mathbf{C}_j$, select the $k$ training examples with the smallest $\bar{d}_j(x_i)$. This approach selects examples that are collectively representative for all members of a cluster, rather than individually optimized per inquiry. The number of clusters $C$ is treated as a hyperparameter, with values 32 or 64 explored in our analysis. Note that this clustering is applied only at evaluation time to guide example selection. This design is intended to improve coverage and diversity of demonstration. Related diversity-preserving subsampling ideas have been studied in other high-dimensional biomedical settings [29].

To study the impact of demonstration count, we vary $k \in \{1, 2, 4, 6, 8, 10, 12, 14, 16\}$ across all selection methods. The learning examples $(x_i, y_i)$ start with a prefix "*Here are demonstration shots*:" Model performance is evaluated for each $k$ and selection method combination across the four entity types. A representative summary of the learning examples and prompt configurations evaluated is presented in **Table 3**.

### 2.4 Task-Level Fine-Tuning

Prompt engineering and in-context learning do not force the model to internalize domain-specific regularities in rare disease NER. We therefore investigate a complementary strategy: task-level fine-tuning. Unlike BioClinicalBERT and BioBERT pretrained on general-domain biomedical corpora, task-level fine-tuning updates the parameters $\boldsymbol{\theta}$ of a pretrained LLM $M_{\boldsymbol{\theta}}$ using the training set of RareDis Corpus, enabling it to learn task-specific patterns rather than relying solely on prompts [30]. For each training pair $(x_i, y_i) \in \mathbf{D}_{\text{train}}$, the prompt $\pi_i$ is constructed by:

$$\pi_i = \text{basic components} \parallel \text{advanced components} \parallel x_i$$

The objective of task-level fine-tuning is to minimize the empirical loss:

$$\min_{\boldsymbol{\theta}} \frac{1}{N} \sum_{i=1}^{N} L(M_{\boldsymbol{\theta}}(\pi_i), y_i)$$

where $L$ is a token-level cross-entropy loss between the identified entity and the ground-truth $y_i$. In this study, we fine-tune the GPT-4o-mini-2024-07-18 model on the RareDis Corpus. Training is conducted using OpenAI's API interface, with hyperparameters batch size, learning rate multiplier, and number of epochs set to "auto". The held-out validation set $\mathbf{D}_{\text{val}}$ is used for early stopping to mitigate overfitting. Training and validation examples are formatted as JSONL records, each containing both the inquiry input $x_i$ and the corresponding entity labels $y_i$, along with the full prompt structure. Unlike in-context learning, no additional examples are prepended at inference time. After fine-tuning, model performance is evaluated on the test set $\mathbf{D}_{\text{test}}$ using the same five-component prompt structure but without any in-context demonstrations.

### 2.5 Retrieval-Augmented Generation Analysis

To augment prompt-based inference with external domain knowledge, we implement a retrieval-augmented generation (RAG) approach in which external reference is dynamically incorporated into the prompt at inference time [31]. This enables the model to access semantically relevant background context without requiring gradient-based parameter updates, contrasting with task-level fine-tuning. We construct a domain-specific knowledge corpus from the Orphanet rare disease alignments database [32]. Alternative biomedical-QA RAG systems have reported only marginal gains when retrieval snippets overlap the prompt content [33]. Let $\mathbf{K_C} = \{(d_i, z_i)\}_{i=1}^{T}$ denote the resulting corpus, where each entry consists of a disease name $d_i$ and corresponding definition snippet $z_i$. The final RAG corpus contains $T = 6{,}860$ entries, each tokenized to a length $\ell_i \in (8, 196)$, with a median of 53 tokens. Each entry $z_i \in \mathbf{K_C}$ is mapped to a semantic embedding $f(z_i) \in \mathbf{R}^{3072}$ using OpenAI's text-embedding-3-large model. Likewise, the inquiry text $x_{\text{inquiry}}$ is embedded as $f(x_{\text{inquiry}})$. We define the retrieval score as the Euclidean distance:

$$d(x_{\text{inquiry}}, z_i) = \left\| f(x_{\text{inquiry}}) - f(z_i) \right\|_2$$

For a given $x_{\text{inquiry}}$, the top-$K$ retrieved knowledge snippets are selected:

$$R(x_{\text{inquiry}}) = \arg \min_{\substack{\mathbf{S} \subset \mathbf{K_C} \\ |\mathbf{S}| = K}} \sum_{z_i \subset \mathbf{S}} d(x_{\text{inquiry}}, z_i)$$

These retrieved snippets are concatenated into a prefix segment of the prompt, "*Here are knowledge snippets*:", followed by the prompts described in previous sections. Two retrieval-augmented prompting strategies are tested:

- **Zero-shot + RAG** – The full prompt consists of only the RAG knowledge prefix and the inquiry input and no labeled learning examples are included:

$$\pi_{\text{zero+RAG}} = \text{basic components} \parallel \text{advanced components} \parallel R(x_{\text{inquiry}}) \parallel x_{\text{inquiry}}$$

- **Few-shot + RAG** – In this setting, $k$ labeled learning examples $\{(x_j, y_j)\}_{j=1}^{k} \subset \mathbf{D}_{\text{train}}$ are included using the **Inquiry-KNN** strategy described in the Section 2.3. The full prompt becomes:

$$\pi_{\text{few+RAG}} = \text{basic components} \parallel \text{advanced components} \parallel$$

$$\{(x_j, y_j)\}_{j=1}^{k} \parallel R(x_{\text{inquiry}}) \parallel x_{\text{inquiry}}$$

We vary $K \in \{1, 2\}$, and $k \in \{1, 2, 4\}$, observing that larger values of $K$ often introduce semantic noise and lead to performance degradation. These two RAG-augmented strategies are evaluated against their non-RAG counterparts to quantify the incremental benefit of incorporating external biomedical knowledge at inference time.

### 2.6 Performance Evaluation Metrics

We formulate rare disease NER as a text-to-entity sequence generation problem, where an LLM outputs a set of entity mentions based on a natural language input. Let the input text be denoted by a token sequence $\{t_1, t_2, \dots, t_n\}$, where $n$ is the total number of tokens. For any given entity type $\tau$, the corresponding ground-truth entity set is given by $\{e_1^{(\tau)}, e_2^{(\tau)}, \dots, e_m^{(\tau)}\}$, where each $e_i^{(\tau)} \in \mathbf{E}^{(\tau)}$, and $\mathbf{E}^{(\tau)}$ is the set of all valid entity strings. The model generates an identified set of entities $\hat{\mathbf{E}}^{(\hat{\tau})} = \{\hat{e}_1^{(\hat{\tau})}, \hat{e}_2^{(\hat{\tau})}, \dots, \hat{e}_k^{(\hat{\tau})}\}$, where $m$ and $k$ may differ.

An entity recognition $\hat{e}_j^{(\hat{\tau})} \in \hat{\mathbf{E}}^{(\hat{\tau})}$ is considered a true positive if there exists a $e_i^{(\tau)} \in \mathbf{E}^{(\tau)}$ such that $\hat{e}_j^{(\hat{\tau})} = e_i^{(\tau)}$ (i.e., exact string match). We denote the number of such correct matches as the true positive for entity type $\tau$:

$$\text{TP}^{(\tau)} = \left|\left\{\hat{e}_j^{(\hat{\tau})} \in \hat{\mathbf{E}}^{(\hat{\tau})} : \hat{e}_j^{(\hat{\tau})} \in \mathbf{E}^{(\tau)}\right\}\right|$$

Accordingly, we define the model evaluation metrics for entity type $\tau$ as follows. Precision is the proportion of identified entities that are correct:

$$\text{Precision}^{(\tau)} = \frac{\text{TP}^{(\tau)}}{|\hat{\mathbf{E}}^{(\tau)}|}$$

Recall is the proportion of ground-truth entities that are correctly identified:

$$\text{Recall}^{(\tau)} = \frac{\text{TP}^{(\tau)}}{|\mathbf{Y}^{(\tau)}|}$$

F1 score is the harmonic mean of precision and recall

$$\text{F1}^{(\tau)} = \frac{2 \times \text{Precision}^{(\tau)}}{\text{Precision}^{(\tau)} + \text{Recall}^{(\tau)}}$$

These metrics are computed separately for each entity type $\tau \in$ {rare disease, disease, sign, symptom} under varying prompt configurations and learning methods described in previous sections.

We construct 95% confidence intervals (CIs) for precision, recall, and F1 using a nonparametric document-level bootstrap on the test set [34]. For each of 2000 bootstrap replicates, we resample test documents with replacement and recompute TP/FP/FN counts and the derived metrics. Reported CIs correspond to the 2.5th and 97.5th percentiles of the bootstrap distribution.

To assess whether retrieval augmentation yields statistically meaningful improvements, we perform a paired and document-level bootstrap on prespecified comparisons. For each bootstrap replicate, we resample the 208 test documents with replacement and recompute document-level precision, recall, and F1 for both the baseline prompt and its matched RAG variant, then record the paired difference $\Delta$ = (RAG - baseline). We report the bootstrap mean $\Delta$ and 95% CI. We additionally report a one-sided bootstrap p-value for improvement. To limit multiple comparisons, we only tested (i) zero-shot + RAG vs zero-shot and (ii) the best performing few-shot setting + RAG vs its few-shot baseline for each entity type.

### 2.7 Error Taxonomy and Quantification

To better understand model behavior, we perform a token-string error analysis on the test set for all four entity types. For each input text $x$, we consider its ground-truth entity set $\mathbf{E}^{(\tau)}$ and the model-identified set $\hat{\mathbf{E}}^{(\hat{\tau})}$ for entity type $\tau$. The recognitions are obtained using **Inquiry-KNN** method, with $k$ selected based on the highest observed F1 score (see **Results** and **Figure 1**). For any input $x$, if no ground-truth entities of type $\tau$ exist, then $\mathbf{E}^{(\tau)} = \emptyset$. Similarly, if the model produces no output for type $\tau$, then $\hat{\mathbf{E}}^{(\hat{\tau})} = \emptyset$. Each identified entity $\hat{e}^{(\hat{\tau})}$ is compared to all ground-truth entities $e^{(\tau)} \in \mathbf{E}^{(\tau)}$ using a case-insensitive token overlap metric:

$$O\left(e^{(\tau)}, \hat{e}^{(\hat{\tau})}\right) = \left|\text{tokens}(e^{(\tau)}) \cap \text{tokens}(\hat{e}^{(\hat{\tau})})\right|$$

A greedy one-to-one alignment procedure is applied, where each recognition is matched to the first available ground-truth span with which it shares the highest token overlap $O\left(e^{(\tau)}, \hat{e}^{(\hat{\tau})}\right) > 0$. Remaining unmatched recognitions and ground-truth entities are retained as spurious and missed, respectively. Aligned entity pairs $\left(e^{(\tau)}, \hat{e}^{(\hat{\tau})}\right)$ are classified into one of six mutually exclusive categories:

- **Correct** – The identified span exactly matches the ground-truth span and the identified entity type matches the true annotation:

$$\hat{e}^{(\hat{\tau})} = e^{(\tau)}, \qquad \hat{\tau} = \tau$$

- **Boundary** – The identified and ground-truth spans have non-zero token overlap but are not identical, with the correct entity type:

$$\hat{e}^{(\hat{\tau})} \neq e^{(\tau)}, \qquad O\left(e^{(\tau)}, \hat{e}^{(\hat{\tau})}\right) > 0, \qquad \hat{\tau} = \tau$$

- **Type** – The identified span exactly matches the ground-truth span, but the entity type is incorrect:

$$\hat{e}^{(\hat{\tau})} = e^{(\tau)}, \qquad \hat{\tau} \neq \tau$$

- **Boundary + Type** – The identified and ground-truth spans overlap but are not identical, and the identified type is incorrect:

$$\hat{e}^{(\hat{\tau})} \neq e^{(\tau)}, \qquad O\left(e^{(\tau)}, \hat{e}^{(\hat{\tau})}\right) > 0, \qquad \hat{\tau} \neq \tau$$

- **Spurious** – The identified entity $\hat{e}^{(\hat{\tau})}$ cannot be aligned to any ground-truth entity of type $\tau$ (i.e., no overlapping span), representing a false positive.
- **Missed** – A ground-truth entity $e^{(\tau)}$ cannot be aligned to any recognition, representing a false negative.

### 2.8 Performance-Cost Analysis

We conduct a performance-cost analysis to quantify how each $k$-shot configuration trades off F1 score against the monetary cost incurred per query using the OpenAI API. Pricing is based on the April 2025 OpenAI pricing sheet, which charges $5 per 1 million input tokens. We compute the number of input tokens for each query under both zero-shot and few-shot settings, ignoring output tokens due to their negligible length in the NER task (typically 10-20 tokens). For each entity type, we compute the average per-query cost for $k = 0, 1, 2, \ldots, 16$. We then regress F1 score against cost to obtain smooth performance-cost curves, aiming to characterize the cost-efficiency of different prompt configurations. Two distinct regression models are adopted based on the empirical shape of the F1-cost relationship for each entity type:

- **Asymptotic-exponential regression.** For entity types exhibiting a monotonic and saturating increase in F1 score (rare disease, disease, and sign), we model the performance-cost curve using a one-phase asymptotic exponential function [35]. We define incremental cost $\Delta c_k = c_k - c_0$ and fit

$$F1(\Delta c) = F_\infty - (F_\infty - F_0)\exp(-\lambda \Delta c),$$

where $F_0$ is the baseline F1 at zero-shot, $F_\infty$ is the asymptotic (plateau) performance, and $\lambda > 0$ controls the saturation rate. We derive the half-rise additional cost $\Delta c_{50} = \log(2)/\lambda$ and the additional cost to achieve 95% of the attainable gain $\Delta c_{95} = \log(20)/\lambda$, and report the corresponding absolute cost as $c_{95} = c_0 + \Delta c_{95}$. Model fitting is performed using nonlinear least squares with a Gauss-Newton optimizer, implemented by `nls` function in R programming language.

- **Local polynomial regression.** The symptom entity does not conform to the monotonic rise assumption; instead, its performance curve is non-monotonic and lacks a well-defined plateau. We therefore conduct local polynomial regression using a Locally Estimated Scatterplot Smoothing (LOESS) method with a span of 0.75 [36]. The smoother fits a first-order local regression at each evaluation point $x$ with tri-cube kernel weighting:

$$w_j(x) = \left(1 - \left|\frac{d_j}{d_{max}}\right|^3\right)^3$$

where $d_j$ is the distance between the evaluation point $x$ and training point $c_j$, and $d_{max}$ is the maximum distance within the local neighborhood defined by the span. The fitted value $\hat{F}(x)$ is obtained by minimizing the locally weighted least squares criterion:

$$\sum_j w_j(x)\{F_j - \beta_0(x) - \beta_1(x)(x - x_j)\}^2$$

where $F_j$ is the observed F1 score at cost value $x_j$. The coefficients $\beta_0(x)$, $\beta_1(x)$ define a locally linear approximation of the performance-cost curve near $x$, with $\hat{F}(x) = \beta_0(x)$ as the locally fitted value. The span hyperparameter (0.75) is selected *a priori* to balance the bias-variance trade-off, given the relatively coarse granularity of the $k$-shot cost grid.

Additionally, we propagate uncertainty to the cost-performance smoothers by applying the same document-level bootstrap at each $k$-shot condition, refitting the entity specific smoother in each replicate, and plotting the pointwise 95% bootstrap bands across cost.

# 3 Results

## 3.1 Zero-Shot Learning

To evaluate model performance in the zero-shot learning setting, we conduct NER analysis using three prompt configurations, each excluding in-context examples. The first design includes only the basic components: task description, output format, and inquiry text. The second design extends this by incorporating task guidance, while the third design adds disambiguation rule on top of the prior components. We compare performance against BioClinicalBERT, a domain-specific BERT model pretrained on PubMed and MIMIC-III corpora and a widely used supervised baseline for clinical NER. In a prior study, Shyr et al. tested BioClinicalBERT on the RareDis Corpus and achieved SOTA performance on rare disease NER [3]. We adopt their reported results as the baseline in this comparison.

**Table 4** summarizes precision, recall, and F1 score across all four entity types for each zero-shot prompt configuration and the BioClinicalBERT baseline. BioClinicalBERT outperforms all zero-shot prompt designs in overall F1, confirming the advantage of supervised learning in this domain. Besides, several trends emerge from the zero-shot results. For rare disease, the basic prompt achieves the highest precision (0.914) across all methods including BioClinicalBERT, though at the cost of reduced recall. Incorporating task guidance and disambiguation rule notably improves recall (from 0.463 to 0.576) and lifts the F1 score from 0.614 to 0.702. This demonstrates that task-specific guidance can significantly enhance recall without severely compromising precision, bringing zero-shot performance closer to the SOTA benchmark (F1 = 0.837).

For disease, a different pattern is observed. The prompt with all components achieves the highest precision (0.545), surpassing BioClinicalBERT (0.494). However, its recall remains low (0.221), leading to a relatively modest F1 score (0.314 vs. 0.491 for SOTA). This suggests that while disambiguation helps filter false positives, it may also suppress valid predictions in this entity type. In the case of sign, the basic prompt yields the highest F1 score (0.392) among zero-shot variants. Task guidance and disambiguation appear to reduce recall (from 0.362 to 0.221), without improving precision. This indicates that the pretrained LLM already captures sufficient contextual cues for sign recognition, and that additional prompt instructions may introduce constraints that hinder generalization. For symptom, precision and F1 score remain low across all configurations. Interestingly, the basic prompt achieves the highest recall (0.653), while the full prompt improves precision (0.142) and F1 score (0.230). Nevertheless, all zero-shot prompts fall well below the BioClinicalBERT (F1 = 0.648), highlighting the difficulty without explicit supervision.

Among the three prompt configurations, the full prompt yields the highest F1 scores for three out of four entity types, suggesting that task-specific definitions and error-aware instructions enhance model performance even without labeled examples. For the sign entity type, the basic prompt alone achieves the best F1 score. This result implies that the LLM pretrained on large-scale general corpora may already encode sufficient knowledge of signs, and that further guidance could introduce redundancy or noise. Overall, while none of the zero-shot prompts match the performance of BioClinicalBERT, they demonstrate competitive precision and recall in certain scenarios, indicating the potential of prompt engineering as a lightweight alternative in low-resource applications.

### 3.2 Few-Shot Learning

**Figure 1** summarizes the F1 scores of few-shot learning across four entity types, comparing different example selection strategies. In general, increasing the number of in-context learning examples $k$ improves model performance, though the degree and pattern of improvement vary by entity type and selection method. For rare disease (**Figure 1A**), all methods except Inquiry-Random show consistent gains as $k$ increases. The Inquiry-KNN strategy consistently outperforms all others and notably exceeds the SOTA (F1 = 0.704) across almost all $k$, with performance peaking around $k = 8$. Interestingly, even a small number of well-selected examples (e.g., $k = 2$) leads to substantial improvement over zero-shot (F1 = 0.702), highlighting the value of semantically aligned demonstrations. A recent multilingual study likewise found that properly selected few-shot cues can outperform fully supervised baselines in English, French, and Spanish clinical NER [19].

For disease (**Figure 1B**), the model again benefits from increased $k$, but the gains plateau earlier, around $k = 4$ to 8. Inquiry-KNN achieves the best results (F1 = 0.518), surpassing the SOTA (F1 = 0.491), and Cluster-KNN follows closely. In contrast, Inquiry-Random yields marginal improvement over zero-shot (F1 = 0.314), underscoring the importance of semantic relevance in example selection. The sign entity (**Figure 1C**) exhibits a slower performance climb, with gains tapering after $k = 8$. Unlike rare disease and disease, Cluster-KNN-64 performs slightly better than Inquiry-KNN across most $k$, suggesting that collective similarity within clusters is more effective than pointwise similarity in this entity type. All three semantic-based methods outperform Inquiry-Random, reinforcing the previous trend. For symptom (**Figure 1D**), performance is underperformed compared to zero-shot (F1 = 0.230) across all methods and values of $k$. In some cases, adding examples degrades performance. This likely reflects the ambiguous nature of symptom annotations, or misalignment between training examples and the model's pretrained representation of medical terms. Among different strategies, Inquiry-KNN yields the highest performance (F1 = 0.223 at $k = 14$).

The low F1 for prompt-based symptom extraction is primarily driven by low precision rather than low recall. In our best few-shot configuration, symptom recall remains moderate (0.673), but precision is very low (0.134), resulting in F1 = 0.223. In contrast, BioClinicalBERT achieves higher and more balanced precision/recall (0.667/0.630), with F1 = 0.648 (**Table 5**). This precision collapse is consistent with our error taxonomy in Section 3.5, where nearly half of symptom outputs are spurious (46%), indicating substantial overgeneration. We also note that symptom is the sparsest entity type in the test set (53 mentions across 208 documents; **Table 1**), which limits coverage for in-context learning and contributes to wider uncertainty.

Comparing selection strategies, we observe that Inquiry-KNN yields the highest F1 scores for rare disease, disease, and symptom, while Cluster-KNN-64 leads on sign. Across entity types, Inquiry-Random consistently underperforms, indicating that semantic similarity – either at the individual or cluster level – is crucial for effective in-context learning. Notably, Cluster-KNN's strong performance demonstrates the potential of collective similarity, which may reduce overfitting to query-specific features that affect pointwise nearest-neighbor strategies like Inquiry-KNN. In addition, Cluster-KNN assigns the same exemplar set to all inputs within a cluster and therefore requires fewer total labeled examples than Inquiry-KNN, which selects a distinct set of examples per query. This makes Cluster-KNN more scalable in scenarios where annotation cost is a bottleneck. Overall, few-shot learning with semantically aligned examples improves performance over zero-shot learning and surpasses SOTA in rare disease and disease entities. These findings underscore the effectiveness of few-shot learning as a low-resource alternative to supervised training when guided by appropriate example selection strategies.

### 3.3 Task-Level Fine-Tuning Achieves Best Overall Performance

**Table 5** compares the NER performance of the task-level fine-tuned GPT-4o-mini model with zero-shot, few-shot, and BioClinicalBERT across four entity types. BioClinicalBERT results are taken from Shyr et al. and included as a reference baseline [3,11]. For zero-shot and few-shot settings, we report the best results across all prompt configurations and learning example selection strategies. Overall, fine-tuning GPT-4o-mini achieves the best performance among the methods evaluated in this study. For rare disease, fine-tuning achieves an F1 score of 0.837, exceeding both the zero-shot (0.702) and few-shot (0.776) variants, and outperforming BioClinicalBERT (0.704), despite its extensive pretraining on general biomedical corpus. The performance gain is especially pronounced in recall (0.822 vs. 0.702 for few-shot), indicating that model exposure to task-specific supervision improves entity coverage substantially. In the disease category, fine-tuning again leads in F1 (0.702), outperforming few-shot (0.518) and BioClinicalBERT (0.491). The fine-tuned model also achieves substantially higher precision (0.713 vs. 0.545) and recall (0.692 vs. 0.511) compared to the few-shot model. For sign, fine-tuning provides the highest F1 score (0.541), benefiting from the strongest recall (0.561) despite slightly lower precision than BioClinicalBERT (0.522 vs. 0.561). This highlights a recall-precision trade-off, where fine-tuning favors completeness over conservatism in recognition.

Compared to other entity types, the symptom exhibits a slightly different trend. BioClinicalBERT achieves the highest F1 score (0.648) and the highest precision (0.667), outperforming the fine-tuned GPT-4o-mini, which attains an F1 of 0.614 with slightly higher recall (0.633 vs. 0.630). The performance gap is narrower here than in other entity types, and both models substantially outperform the zero-shot (F1 = 0.230) and few-shot (F1 = 0.223) models. These results suggest that the more ambiguous and context-sensitive symptom extraction benefits from broad biomedical pretraining and may require additional contextual reasoning beyond prompt-based learning.

Across all four entity types, task-level fine-tuned GPT-4o-mini consistently yields balanced precision and recall, indicating robust generalization and reliability for NER tasks. In contrast, zero-shot prompting tends to favor precision at the expense of recall, while few-shot learning offers intermediate gains but does not consistently close the performance gap, particularly on high-recall tasks. These findings reinforce the value of full-model fine-tuning when task-specific labeled data is available. Unlike BioClinicalBERT, which is trained on general-purpose biomedical corpora (PubMed, MIMIC-III), GPT-4o-mini benefits from being directly fine-tuned on the RareDis Corpus, allowing it to internalize domain-specific terminology, entity structure, and annotation conventions. However, it is important to acknowledge the resource-intensive nature of fine-tuning, which requires substantial annotation effort, model retraining, and validation infrastructure. In contrast, few-shot learning achieves near-SOTA results at a fraction of the labeling cost, particularly for rare disease and sign.

### 3.4 Impact of RAG on Zero- and Few-Shot Learning

**Table 6** presents the performance gains achieved by augmenting prompts with one or two knowledge snippets retrieved from the RAG corpus, across zero-shot and few-shot learning settings (with 1, 2, or 4 learning examples). Only metric-entity combinations that show improvement with RAG augmentation are reported. Among the 48 metric-entity combinations evaluated, 13 (27%) show higher point estimates with RAG augmentation. The degree of benefit varies across evaluation metrics: 7 out of 16 (44%) precision scores improve, compared to 3 out of 16 (19%) for recall and 3 out of 16 (19%) for F1 score. The average gains are modest, with 0.016 for precision, 0.045 for recall, and 0.013 for F1 score. By entity type, symptom recognition shows the greatest relative benefit (6 of 12 metrics improved, 50%), followed by sign (3 of 12, 25%), disease (2 of 12, 13%), and rare disease (2 of 12, 13%).

We further evaluate whether these changes are statistically supported using paired and document-level bootstrap tests on prespecified comparisons (**Supplementary Table S1**). Across most entities and settings, RAG does not yield a statistically significant improvement in F1, and several comparisons showed negative ΔF1. The clearest benefit is for sign extraction in the zero-shot setting**,** where adding one retrieved snippet improves F1 by 0.030 (95% CI (0.005,0.055)). In contrast, for rare disease zero-shot using two snippets, F1 decreases by 0.108 (95% CI (−0.152, −0.064)), and for symptom, zero-shot RAG increases recall but reduces precision such that F1 declines overall by −0.041 (95% CI (−0.083, −0.001)).

These results suggest that, in the context of rare disease NER, RAG provides limited additional value when high-quality prompts and relevant learning examples are already available. Improvements in precision imply that RAG snippets may help suppress false positives in select cases. However, the gains are marginal and do not scale with the number of snippets included. Recall also shows small improvement, except for modest boosts in sign and symptom recognition, indicating that RAG does not substantially surface new entities beyond what is already captured by the base prompt. Given that each RAG snippet contains roughly 50 tokens and requires a separate embedding computation, the cost-benefit trade-off becomes unfavorable for scenarios constrained by token budget or inference latency. In such settings, allocating resources toward additional learning examples or lightweight fine-tuning may yield better returns.

Two factors likely explain the limited benefit of RAG in this task. First, GPT-4o model already encodes substantial biomedical knowledge, including lexical variants and factual associations relevant to rare and common diseases. As such, short RAG snippets often add little new information. Second, overlap between few-shot learning examples and retrieved snippets, in terms of disease mentions and syntactic structure, further diminishes the incremental value of RAG. When retrieved content redundantly mirrors patterns already shown in the prompt, its utility drops to near zero. RAG's most consistent benefit is recall improvement for the most challenging entity types. For example, in the zero-shot setting, adding one retrieved snippet increases sign recall from 0.221 to 0.254 and symptom recall from 0.612 to 0.673 (**Table 6**). For symptoms, this recall increase can coincide with reduced precision, so overall F1 may not increase even when recall improves.

### 3.5 Error Analysis

**Figure 2** visualize model-identified entities into six mutually exclusive categories defined in Section 2.7, providing a fine-grained view of model behavior. Rare disease recognition emerges as the most robust, with over half of all mentions correctly identified with exact span and type agreement (Correct = 51%), roughly twice the rate observed for disease (30%) and sign (23%). Most remaining errors are relatively benign, comprising modest span deviations (Boundary = 20%) and low-severity false positives (Spurious = 15%). Omissions are comparatively infrequent (Missed = 14%), and the near absence of entity type confusion (Type + Boundary and Type < 1%) confirms that rare disease are rarely mislabeled.

The disease and sign categories exhibit complementary error profiles. Disease recognition is primarily limited by recall, with nearly one-third of ground-truth mentions undetected (Missed = 29%). Sign recognition, on the other hand, suffers more from low precision: boundary drift affects 31% of recognitions, and a further 33% are spurious hallucinations, indicating substantial over-generation. Together, these two entity types account for a substantial portion of the overall error volume illustrated in **Figure 2** and highlight the divergent sources of model failure.

The symptom category presents a distinct pattern. While the model identifies a relatively large proportion of entities (Correct = 35%), nearly half of all outputs are unsupported by the ground truth (Spurious = 46%). To quantify whether these false positive symptom outputs reflect sign/symptom boundary ambiguity, we compare spurious symptom strings against sign annotations in the same documents under zero-shot setting. Among 182 spurious symptom outputs, 101 (55.5%) exactly match a sign mention, while the remaining 81 (44.5%) are unsupported by either symptom or sign annotations. This suggests symptom errors are driven by both sign/symptom boundary confusion and unsupported overgeneration. The remediation may require refining entity boundaries or prompt specificity than on increasing the number of examples.

Overall, Boundary, Spurious, and Missed collectively account for the majority of errors, with Spurious alone exceeding 30% in Sign and 45% in Symptom. This indicates that improvements in model performance may be more effectively achieved through post-processing heuristics, such as dictionary-based filtering to suppress unsupported outputs and head-noun alignment to correct span drift. From a deployment standpoint, these error profiles further motivate a human-in-the-loop use case, where extracted entities are treated as candidates requiring expert verification.

### 3.6 Performance-Cost Trade-off

**Figure 3** illustrates the F1 scores achieved with zero- and few-shot learning as a function of the corresponding per-query cost. The asymptotic-exponential model provides a close fit to the observed points for rare disease, disease, and sign, as indicated by low RMSE and high pseudo-$R^2$ (**Supplementary Table S2**) and by residual plots showing no systematic patterns across cost (**Supplementary Figure S2**). Symptom remains non-monotonic, so we do not interpret plateau-based parameters for symptom. For rare disease, disease, and sign entities, the F1-cost relationship exhibits a smooth saturating trend, well-modeled by an asymptotic-exponential function. In contrast, the symptom entity displays pronounced non-monotonic behavior, for which a LOESS smoother is used. The confidence bands in **Figure 3** also indicate wider uncertainty for symptom than for other entities, consistent with its small number of test mentions (**Table 1**). The estimated performance ceiling $F_\infty$ and the half-rise cost $c_{0.5} = \ln 2/\lambda$ for the exponential fits are summarized in **Table 7**.

For rare disease recognition, F1 score increases from 0.702 at zero-shot (0.19¢) to 0.760 at $k = 4$ (0.64¢), reaching 96% of the estimated ceiling. Beyond this point, each additional cent of inference cost contributes less than 0.003 F1 improvement, showing rapid saturation. For disease and sign, performance plateaues near F1 ≈ 0.50, with half of the total gain achieved at a cost of approximately 0.10¢ ($k = 2$). However, an additional 0.9¢ is required to close the final 5% of the performance gap ($k = 8$). Past this threshold, marginal returns drop below 0.002 F1 per additional cent. For symptom recognition, the fitted LOESS curve fluctuates within $\pm 0.021$ of the baseline F1 ≈ 0.230 across the entire 0–2¢ cost range, revealing no meaningful benefit from increasing the number of learning examples. This result suggests that token budget is largely ineffective for improving model accuracy on this entity type.

Overall, a uniform budget cap of ~1¢ per query, equivalent to up to 8-shot prompting, is sufficient to capture ≥ 95% of the attainable performance for rare disease, disease, and sign entities. Beyond this threshold, further spending results in negligible gains. For symptoms, improvements are likely better achieved through alternative strategies, such as data augmentation, label refinement, or fine-tuning, rather than through prompt expansion.

## 4 Discussion

This study demonstrates that prompt-based LLMs can achieve competitive performance in rare disease NER without extensive task-specific training data. One key finding is the importance of prompt design and learning example selection. We observe that semantic selection of in-context examples consistently improves NER accuracy over random example selection. This is in line with recent reports that contextually relevant demonstrations boost medical NER performance [21]. Incorporating external knowledge via RAG yields modest average gains overall, consistent with GPT-4o already possessing substantial biomedical knowledge. However, RAG can still be useful as a targeted recall booster for challenging entities such as signs and symptoms. This result diverges from the large improvements RAG has shown on knowledge-intensive QA tasks [20], indicating that for rare disease NER, the bottleneck is less about world knowledge but more about recognizing precise spans in context. Our analysis also highlights the cost-efficiency of the prompt-based approach. With only a handful of well-chosen examples, GPT-4o achieves strong results at a fraction of the total cost than collecting and curating a large expert-annotated corpus. In essence, prompt-based GPT-4o offers high returns for low investment, making it an attractive solution in low-resource NER scenarios [37].

Our NER pipeline represents a departure from traditional supervised approaches in biomedical NER. Historically, state-of-the-art results come from language models like BioBERT and BioClinicalBERT pretrained on general biomedical corpora, or earlier from statistical sequence taggers and LSTM-based models [38]. Recent studies begin to explore the potential of LLM in using prompt engineering. For instance, Agrawal et al. show that GPT-3 could perform few-shot clinical information extraction comparably to fully trained models [39]; Xi et al. apply GPT-based methods to Reddit posts to characterize patient-reported manifestations of sarcoidosis [40]. The significance of our findings is underscored by comparisons to earlier work on rare disease text mining, which are often bottlenecked by data scarcity [41]. We confirm these observations and demonstrate that a next-generation LLM can substantially close the performance gap with domain-trained models. Our work builds upon and goes beyond prior insights, showing that prompt-based LLMs can achieve near-parity with SOTA in low-resource rare disease NER.

The performance of GPT-4o in the few-shot learning context suggests that institutions can leverage a pretrained LLM to perform entity recognitions without large-scale data annotation. In settings where rapid deployment is valued over absolute peak performance, our prompt-based pipeline offers a compelling solution. LLM powered NER system can also be integrated into electronic health records to surface candidate rare disease mentions in physicians' notes and support retrospective cohort screening, with expert verification before downstream clinical use. Another use case is biomedical knowledge curation, in which researchers could use GPT-4o to extract disease-phenotype associations from research papers or case reports. The system can also be easily repurposed for new subtasks by modifying the prompts, rather than retraining models. Because the NER logic resides in the prompt and model rather than custom code, deploying the system can be as simple as calling an API endpoint. This lowers the barrier for institutions that lack extensive machine learning infrastructure.

Our error taxonomy highlights that the dominant failure modes vary by entity type. In particular, spurious outputs remain substantial for sign and symptom, which in a clinical setting could translate into false alerts or unnecessary downstream review. Conversely, missed entities reduce sensitivity and may limit utility for automated cohort identification if used without safeguards, while boundary errors can impede normalization to controlled vocabularies and reduce interoperability with structured clinical systems. Taken together, these patterns indicate that the proposed pipeline is best positioned as a screening aid that surfaces candidate entities for expert confirmation, rather than as an autonomous diagnostic tool. In practice, the most appropriate workflow is that model outputs should be reviewed prior to any clinical interpretation.

The symptom results also highlight a broader limitation of prompt-only extraction for clinically nuanced categories. Prompt-based methods tend to overgenerate plausible clinical complaints, producing many false positives and depressing precision and F1. This is consistent with the high Spurious rate in **Figure 2**. At the same time, the strong symptom performance of supervised models of BioClinicalBERT and the substantial improvement under lightweight fine-tuning suggest that task specific supervision is important for learning the corpus specific symptom boundary, especially when the test set contains relatively few symptom mentions (**Table 1**).

From a practical standpoint, the symptom errors suggest several low cost mitigation strategies that directly target the dominant failure modes. First, because a substantial fraction of spurious symptom strings reflects sign/symptom boundary confusion (Section 3.5), prompts can be made more corpus-aware by explicitly restating the RareDis symptom definition and adding a small set of negative examples that are objective findings to discourage type swaps. Second, spurious symptom outputs can be reduced via lightweight post-processing, such as dictionary filtering to suppress unsupported generic complaints and a consistency check that removes or relabels symptom outputs that match a sign in the same document. Finally, missed symptoms may be addressed by providing more targeted in-context guidance for this nuanced category. It is helpful to retrieve short usage examples of symptom mentions rather than generic definitions or use a two-stage candidate generation plus verification prompt. These hybrid prompt and post-processing strategies provide a practical path to improving symptom extraction without requiring full model retraining.

Our study has several limitations that warrant discussion. First, prompting still lags behind specialized models in certain NER tasks. These gaps are consistent with recent evaluations showing that general purpose LLMs can underperform fine-tuned domain models on token-level clinical NER under strict boundary matching, even with prompt engineering or few-shot prompting [18,19,42,43]. For scenarios requiring strict annotation fidelity, post-processing or alignment is needed to refine the raw outputs of the LLM. The second limitation is the marginal benefit observed from RAG in our experiments. One possible reason is that GPT-4o already encodes a wealth of medical knowledge from its pretraining, and additional snippets contribute limited new information. It is also plausible that our retrieval method does not select sufficiently targeted context. More sophisticated retrieval, such as grabbing example sentences of the exact entity usage, might yield a greater benefit [21]. We outline a concrete implementation and evaluation plan as future work below. Finally, the reliance on API is a practical limitation. Using a closed-source model means that reproducibility and long-term deployment are not fully guaranteed. Sending sensitive patient data to an external API can also conflict with privacy regulations [44]. Therefore, deploying a similar system in a hospital setting would require robust de-identification solutions [45,46].

There are several avenues to extend current study. One direction is to combine the strengths of LLMs with rule-based systems. After generating candidate entities, a post-processing step could apply heuristic rules or dictionary matching to correct span boundaries and unify terminology [23]. Even simple alignment rules, such as ensuring the output exactly matches a known rare disease name, could substantially increase precision without requiring model retraining. Another promising avenue is self-consistency decoding, which generates multiple outputs for the same input and then taking a majority vote among the answers [47]. Similarly, incorporating chain-of-thought prompting may help the model internally reason about the text [48]. By guiding the model through intermediate reasoning steps, it is possible to resolve ambiguities and improve the recognition of difficult entities. Lastly, a hybrid strategy worth exploring is to fine-tune a model on synthetic annotations generated by the LLM. Recent work suggests that LLMs can create high-fidelity synthetic data for training downstream models [49]. To apply this strategy, GPT-4o could annotate a large collection of unlabeled clinical texts, possibly with iterative refinement or human review. A compact model fine-tuned on this corpus might then serve as a cost-effective and privacy-preserving solution that approaches SOTA.

A particularly promising direction is to make retrieval more targeted than the definition style snippets used in our current RAG setting. One practical approach is entity centric and sentence-level retrieval. We would index short sentences rather than whole documents from a reference corpus and attach lightweight metadata such as the entity type and the entity strings occurring in the sentence. At inference time, instead of retrieving generic descriptions, we would retrieve a small set of usage examples that (i) are semantically similar to the local context in the input text and (ii) contain the same or closely matched entity strings, so the retrieved evidence demonstrates how the entity appears in natural clinical language and how boundaries are annotated. These retrieved usage sentences could be injected into the prompt as concise demonstrations to reduce boundary errors and type confusions. For larger corpora, more computationally efficient subsampling methods could also be explored to select representative candidate demonstrations under a fixed token budget [50–52]. We would evaluate this targeted strategy via an ablation against our current RAG approach under a matched token budget, reporting performance deltas with bootstrap confidence intervals and examining shifts in the error taxonomy to identify which error modes are most affected.

Complementary to prompt centric directions, advances in supervised deep learning for sequential prediction remain relevant for biomedical NER. Traditional NER is commonly formulated as token-level sequence labeling, where performance can depend strongly on sequence modeling choices and training configuration. Recent work in imbalanced sequential settings shows that systematic hyperparameter optimization and targeted data augmentation can materially improve performance when minority class examples are scarce, and performance can be sensitive to training configuration and hyperparameter choices [53]. For example, Shukla et al. combine a temporal convolutional sequence model with GAN based minority class augmentation and automated hyperparameter optimization in an imbalanced setting [54]. Analogous approaches such as structured decoding for sequence labeling, automated hyperparameter searches for supervised baselines, and augmentation of rare entity mentions are promising complements to prompt based pipelines and may further mitigate the long-tail distribution challenges typical of rare disease corpora. In our setting, these imbalance-aware approaches would directly target low frequency entities such as symptoms and ultra rare disease mentions, where sparse supervision can lead to unstable performance and wider uncertainty.

In conclusion, our study shows that thoughtful prompt engineering and use of learning examples can serve as a powerful tool for rare disease NER. We have discussed how our findings both align with and extend prior knowledge, the practical trade-offs involved, and the limitations that temper the results. By addressing those limitations through the future directions outlined above, we anticipate that prompt-based LLM approaches will become even more accurate, interpretable, and integrated into real-world biomedical text mining pipelines.

## Conflicts of Interest

Nan Miles Xi and Yu Deng are full-time employees of AbbVie. AbbVie has no role in the design, analysis, interpretation, or decision to publish this study. All other authors declare no conflicts of interest.

## Data Availability

The RareDis Corpus analyzed in this study is publicly available from the NLP4RARE-CM-UC3M repository at https://github.com/isegura/NLP4RARE-CM-UC3M. The same data in rds format can also be downloaded from Zenodo repository at https://zenodo.org/records/18011070.

The Orphanet data used to construct the retrieval corpus for the RAG experiments are publicly available from Orphadata at https://www.orphadata.com/orphanet-scientific-knowledge/. A clean csv version can also be downloaded from Zenodo repository at https://zenodo.org/records/18011070.

All code needed to reproduce the results of this study is openly accessible at GitHub repository https://github.com/xnnba1984/Leveraging-Large-Language-Models-for-Rare-Disease-Named-Entity-Recognition/tree/main.

## Funding

Lin Wang is supported by the National Science Foundation (DMS-2413741) and the Central Indiana Corporate Partnership AnalytiXIN Initiative.

## Author Contributions

Nan Miles Xi performed prompt engineering framework, performance evaluation, and manuscript writing. Yu Deng contributed to data preprocessing, embedding generation, model implementation, and results analysis. Lin Wang led the conceptualization and study design, provided guidance on statistical analysis and manuscript review. All authors reviewed and approved the final version of the manuscript.

# Tables and Figures

**Table 1. Summary statistics of RareDis Corpus dataset and definitions of its named entities.**

<table>
<tr><th></th><th></th><th>Training set</th><th>Validation set</th><th>Test set</th><th>Total</th></tr>
<tr><td colspan="2">Number of documents</td><td>729</td><td>104</td><td>208</td><td>1,041</td></tr>
<tr><td rowspan="4"><b>Named entity</b></td><td>Rare disease</td><td>3,608</td><td>525</td><td>1,088</td><td>5,221</td></tr>
<tr><td>Disease</td><td>1,647</td><td>230</td><td>471</td><td>2,348</td></tr>
<tr><td>Sign</td><td>3,744</td><td>528</td><td>1,061</td><td>5,333</td></tr>
<tr><td>Symptom</td><td>319</td><td>24</td><td>53</td><td>396</td></tr>
<tr><td rowspan="4"><b>Definition</b></td><td>Rare disease</td><td colspan="4">A rare disease is a health condition that affects a small percentage of the population. In the U.S., a disease is considered rare if it affects fewer than 200,000 people. In the European Union, a disease is considered rare if it affects fewer than 1 in 2,000 people.</td></tr>
<tr><td>Disease</td><td colspan="4">A disease is a condition of the body or mind that impairs normal functioning and is characterized by specific signs and symptoms. Diseases can be caused by a variety of factors, including infections, genetic mutations, environmental factors, and lifestyle choices</td></tr>
<tr><td>Sign</td><td colspan="4">A sign of a disease is objective evidence of disease that can be observed or detected by someone other than the individual affected by the disease. It includes measurable indicators such as physical findings, laboratory test results, and imaging studies, which provide concrete evidence of a medical condition.</td></tr>
<tr><td>Symptom</td><td colspan="4">A symptom is the subjective experience reported by the patient, which cannot be directly observed or measured by others. They reflect what the patient feels, such as pain, fatigue, or nausea. Symptoms are experienced internally and rely on the patient’s description.</td></tr>
</table>

**Table 2. Prompt design components and task-specific instructions used for extracting each entity type.**

| Prompt component | Content |
|---|---|
| Task description | *Identify the names of (entity) from the following text.* |
| Output format | *Output only the exact (entity) names without any additional changes. If there are multiple (entity), separate their names with commas. If there is no (entity), output 'none'.* |
| Input text | *The text from which you need to extract the names of (entity) is …* |
| Task guidance | **Rare disease:** *A rare disease is a health condition that affects a small percentage of the population. In the U.S., a disease is considered rare if it affects fewer than 200,000 people. In European Union, a disease is considered rare if it affects fewer than 1 in 2,000 people.*<br>**Disease:** *A disease is a condition of the body or mind that impairs normal functioning and is characterized by specific signs and symptoms. Diseases can be caused by a variety of factors, including infections, genetic mutations, environmental factors, and lifestyle choices.*<br>**Sign:** *A sign of a disease is the objective evidence of disease that can be observed or detected by someone other than the individual affected by the disease. It includes measurable indicators such as physical findings, laboratory test results, and imaging studies, which provide concrete evidence of a medical condition.*<br>**Symptom:** *A symptom is the subjective experience reported by the patient, which cannot be directly observed or measured by others. They reflect what the patient feels, such as pain, fatigue, or nausea. Symptoms are experienced internally and rely on the patient's description.* |
| Disambiguation rule | **Rare disease:** *Treat abbreviations as separate rare disease names. Do not identify regular diseases as rare diseases.*<br>**Disease:** *Differentiate between rare diseases and diseases. A rare disease is a health condition that affects a small percentage of the population. Rare diseases are a subset of diseases. Only output diseases, not rare diseases.*<br>**Sign:** *Differentiate between signs and symptoms. Symptoms are subjective experiences of disease reported by the patient and cannot be directly measured by healthcare providers. Only output signs, not symptoms.*<br>**Symptom:** *Differentiate between symptoms and signs. Signs are objective indicators of a disease that can be observed, measured, or detected by someone other than the patient, such as a doctor or medical professional. Only output symptoms, not signs.* |

**Table 3. Exemplary in-context learning examples contained in the prompts.**

| Named entity | Content |
| --- | --- |
| Rare disease | *Input text: Myhre syndrome is an extremely rare inherited disorder that, in theory, affects males and females in equal numbers. More than 60 cases have been reported in medical literature. Because some cases of Myhre syndrome most likely go undiagnosed or misdiagnosed, determining the true frequency of the disorder in the general population is difficult.*<br><br>*Output: myhre syndrome.* |
| Disease | *Input text: May-Hegglin Anomaly is a rare, inherited, blood platelet disorder characterized by abnormally large and misshapen platelets (giant platelets) and defects of the white blood cells known as leukocytes. The defect of the white blood cells consists of the presence of very small (2-5 micrometers) rods, known as Dohle bodies, in the fluid portion of the cell (cytoplasm). Some people with this disorder may have no symptoms while others may have various bleeding abnormalities. In mild cases, treatment for May-Hegglin Anomaly is not usually necessary. In more severe cases, transfusions of blood platelets may be necessary. May-Hegglin Anomaly is a rare blood platelet disorder that affects males and females in equal numbers. It occurs more often in people of Greek or Italian descent than among others. As of about 10 years ago, only about 170 cases were reported in the literature.*<br><br>*Output: inherited, blood platelet disorder, blood platelet disorder.* |
| Sign | *Input text: The autonomic nervous system controls involuntary actions such as widening or narrowing of our blood vessels. Failure in this system can lead to orthostatic hypotension, which means a sudden drastic drop in blood pressure especially from a lying or sitting down position. The exact cause of pure autonomic failure (PAF) is not known, but is defined as autonomic failure without central nervous system (brain or spinal cord) involvement. PAF is caused by abnormal accumulation of a protein called alpha-synuclein in autonomic nerves. This protein helps nerve cells communicate, but its function is not fully understood. Patients with PAF have a loss of nerve cells (neurons) in the intermediolateral column of the spinal cord. The worldwide prevalence of PAF is not known. The age of onset is during adulthood usually in individuals over 60 years. It is more common in males than in females.*<br><br>*Output: orthostatic hypotension, sudden drastic drop in blood pressure, accumulation of a protein called alpha-synuclein in autonomic nerves.* |
| Symptom | *Input text: Carbamoyl phosphate synthetase I deficiency (CPSID) is a rare inherited disorder characterized by complete or partial lack of the carbamoyl phosphate synthetase (CPS) enzyme. This is one of five enzymes that play a role in the breakdown and removal of nitrogen from the body, a process known as the urea cycle. The lack of the CPSI enzyme results in excessive accumulation of nitrogen, in the form of ammonia (hyperammonemia), in the blood. Affected children may experience vomiting, refusal to eat, progressive lethargy, and coma. CPSID is inherited as an autosomal recessive genetic disorder. The estimated frequency of CPSID is 1 in 150-200,000 births. The estimated frequency of urea cycle disorders collectively is one in 30,000. However, because urea cycle disorders like CPSID often go unrecognized, these disorders are under-diagnosed, making it difficult to determine the true frequency of urea cycle disorders in the general population.*<br><br>*Output: refusal to eat, progressive lethargy.* |

**Table 4. NER performance of different prompt designs under zero-shot learning.** The BioClinicalBERT baseline is taken from Shyr et al., who fine-tuned BioClinicalBERT on the RareDis Corpus [3,11]. The best performances among different prompt designs and BioClinicalBERT model are underscored for each task. The 95% CIs constructed by bootstrap are shown in parentheses. CIs are computed for GPT based experiments and not available for the BioClinicalBERT baseline reported in the literature.

| Named entity | Prompt and model | Precision | Recall | F1 score |
|---|---|---|---|---|
| Rare disease | Basic | **0.914**<br>(0.870, 0.952) | 0.463<br>(0.419, 0.512) | 0.614<br>(0.569, 0.661) |
| | Basic + Task guidance | 0.873<br>(0.799, 0.935) | 0.442<br>(0.395, 0.498) | 0.587<br>(0.537, 0.640) |
| | Basic + Task Guidance + Disambiguation rule | 0.897<br>(0.842, 0.944) | 0.576<br>(0.525 0.633) | 0.702<br>(0.657, 0.749) |
| | BioClinicalBERT | 0.689 | **0.720** | **0.704** |
| Disease | Basic | 0.230<br>(0.174, 0.287) | 0.282<br>(0.210, 0.353) | 0.253<br>(0.191, 0.314) |
| | Basic + Task guidance | 0.252<br>(0.193, 0.311) | 0.297<br>(0.229, 0.361) | 0.273<br>(0.211, 0.330) |
| | Basic + Task Guidance + Disambiguation rule | **0.545**<br>(0.448, 0.642) | 0.221<br>(0.169, 0.274) | 0.314<br>(0.249, 0.376) |
| | BioClinicalBERT | 0.494 | **0.488** | **0.491** |
| Sign | Basic | 0.426<br>(0.379, 0.475) | 0.362<br>(0.315, 0.411) | 0.392<br>(0.347, 0.437) |
| | Basic + Task guidance | 0.387<br>(0.328, 0.444) | 0.257<br>(0.214, 0.305) | 0.309<br>(0.261, 0.358) |
| | Basic + Task Guidance + Disambiguation rule | 0.377<br>(0.323, 0.429) | 0.221<br>(0.187, 0.265) | 0.278<br>(0.238, 0.327) |
| | BioClinicalBERT | **0.561** | **0.516** | **0.538** |
| Symptom | Basic | 0.048<br>(0.028, 0.072) | **0.653**<br>(0.489, 0.806) | 0.090<br>(0.053, 0.131) |
| | Basic + Task guidance | 0.097<br>(0.058, 0.144) | 0.592<br>(0.415, 0.775) | 0.167<br>(0.102, 0.237) |
| | Basic + Task Guidance + Disambiguation rule | 0.142<br>(0.087, 0.207) | 0.612<br>(0.452, 0.780) | 0.230<br>(0.146, 0.316) |
| | BioClinicalBERT | **0.667** | 0.630 | **0.648** |

**Table 5. NER performance of different models and learning methods.** Each metric in zero-shot and few-shot learning is the best result across all prompt-example configurations. The best performances among different models are underscored for each task. The 95% CIs constructed by bootstrap are shown in parentheses. CIs are computed for GPT based experiments and not available for the BioClinicalBERT baseline reported by Shyr et al. [3].

| Named entity | Model | Precision | Recall | F1 score |
|---|---|---|---|---|
| Rare disease | Fine-tuning | 0.853<br>(0.809, 0.898) | **0.822**<br>(0.783, 0.859) | **0.837**<br>(0.805, 0.870) |
| | Zero-shot | 0.914<br>(0.870, 0.952) | 0.576<br>(0.525 0.633) | 0.702<br>(0.657, 0.749) |
| | Few-shot | **0.920**<br>(0.881, 0.955) | 0.702<br>(0.653, 0.752) | 0.776<br>(0.735, 0.814) |
| | BioClinicalBERT | 0.689 | 0.720 | 0.704 |
| Disease | Fine-tuning | **0.713**<br>(0.659, 0.765) | **0.692**<br>(0.636, 0.748) | **0.702**<br>(0.658 0.745) |
| | Zero-shot | 0.545<br>(0.448, 0.642) | 0.297<br>(0.229, 0.361) | 0.314<br>(0.249, 0.376) |
| | Few-shot | 0.545<br>(0.466, 0.622) | 0.511<br>(0.432, 0.549) | 0.518<br>(0.456, 0.567) |
| | BioClinicalBERT | 0.494 | 0.488 | 0.491 |
| Sign | Fine-tuning | 0.522<br>(0.480, 0.562) | **0.561**<br>(0.515, 0.605) | **0.541**<br>(0.501, 0.579) |
| | Zero-shot | 0.426<br>(0.379, 0.475) | 0.362<br>(0.315, 0.411) | 0.392<br>(0.347, 0.437) |
| | Few-shot | 0.463<br>(0.418, 0.504) | 0.494<br>(0.444, 0.538) | 0.478<br>(0.432, 0.517) |
| | BioClinicalBERT | **0.561** | 0.516 | 0.538 |
| Symptom | Fine-tuning | 0.596<br>(0.458, 0.767) | 0.633<br>(0.492, 0.769) | 0.614<br>(0.500, 0.738) |
| | Zero-shot | 0.142<br>(0.087, 0.207) | 0.612<br>(0.452, 0.780) | 0.230<br>(0.146, 0.316) |
| | Few-shot | 0.134<br>(0.079, 0.198) | **0.673**<br>(0.508, 0.833) | 0.223<br>(0.141, 0.309) |
| | BioClinicalBERT | **0.667** | 0.630 | **0.648** |

**Table 6. NER performance gains from RAG relative to in-context learning.** A $k$-shot model refers to one prompted with $k$ labeled learning examples selected using the Inquiry-KNN method. The "+ $n$-RAG" condition additionally prepends $n$ knowledge snippets retrieved from the RAG corpus. Only metrics showing performance improvement with RAG snippets are reported. A dash (-) indicates no observed benefit. The best-performing configuration for each task is underscored. The 95% confidence intervals constructed by bootstrap are shown in parentheses.

| Named entity | Model | Precision | Recall | F1 score |
|---|---|---|---|---|
| Rare disease | 2-shot | 0.870<br>(0.824, 0.913) | - | - |
| | 2-shot + 1-RAG | **0.871**<br>(0.829, 0.910) | - | - |
| | 4-shot | 0.855<br>(0.807, 0.902) | - | - |
| | 4-shot + 2-RAG | **0.886**<br>(0.846, 0.924) | - | - |
| Disease | 2-shot | 0.525<br>(0.453, 0.599) | - | - |
| | 2-shot + 2-RAG | **0.545**<br>(0.466, 0.62) | - | - |
| | 4-shot | 0.508<br>(0.434, 0.579) | - | - |
| | 4-shot + 2-RAG | **0.534**<br>(0.460, 0.607) | - | - |
| Sign | Zero-shot | 0.377<br>(0.323, 0.429) | 0.221<br>(0.187, 0.265) | 0.278<br>(0.238, 0.327) |
| | Zero-shot + 1-RAG | **0.405**<br>(0.344, 0.467) | **0.254**<br>(0.210, 0.300) | **0.312**<br>(0.263, 0.360) |
| Symptom | Zero-shot | - | 0.612<br>(0.452, 0.780) | - |
| | Zero-shot + 1-RAG | - | **0.673**<br>(0.515, 0.824) | - |
| | 2-shot | 0.114<br>(0.067, 0.163) | - | 0.193<br>(0.118, 0.264) |
| | 2-shot + 2-RAG | **0.118**<br>(0.069, 0.172) | - | **0.197**<br>(0.119, 0.277) |
| | 4-shot | 0.115<br>(0.068, 0.170) | 0.612<br>(0.440, 0.786) | 0.194<br>(0.120, 0.275) |
| | 4-shot + 1-RAG | **0.117**<br>(0.068, 0.173) | **0.653**<br>(0.500, 0.803) | **0.198**<br>(0.121, 0.279) |

**Table 7. Asymptotic performance and cost-efficiency metrics across four named entity types.** For rare disease, disease, and sign, plateau and cost-efficiency metrics are derived from the fitted asymptotic-exponential model. $c_{95}$ denotes the per-query cost required to reach 95% of the attainable gain relative to zero-shot.

| Named entity | Plateau $F_\infty$ | Half-rise cost $c_{0.5}$ | Cost to reach 95% of attainable gain | Description |
|---|---|---|---|---|
| Rare disease | 0.763 | 0.07 ¢ | 0.62 ¢ ($k \approx 4$) | Fastest and highest saturation |
| Disease | 0.495 | 0.11 ¢ | 1.05 ¢ ($k \approx 8$) | Gains diminish beyond 8-shot |
| Sign | 0.465 | 0.08 ¢ | 0.94 ¢ ($k \approx 8$) | Mirrors disease trend |
| Symptom | 0.230 | — | — | No systematic cost response |

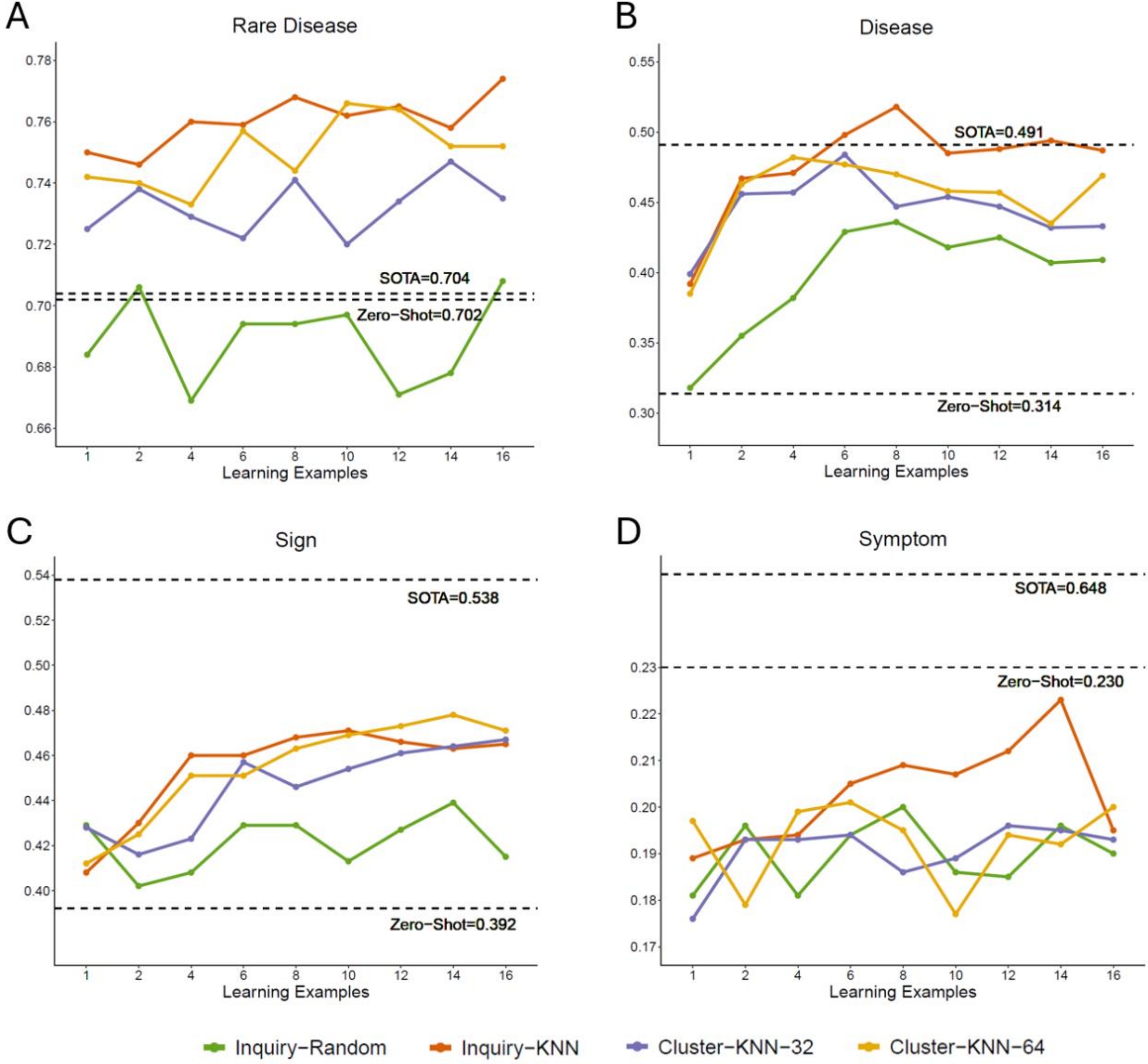

**Figure 1. F1 scores of few-shot learning across different numbers of learning examples. A.** Rear disease. **B.** Disease. **C.** Sign. **D.** Symptom. Two dash lines indicate the state-of-the-art (SOTA) performances using BioClinicalBERT model reported by Shyr et al. [3] and the best performance of prompt designs without learning examples (zero-shot) reported in **Table 4,** respectively.

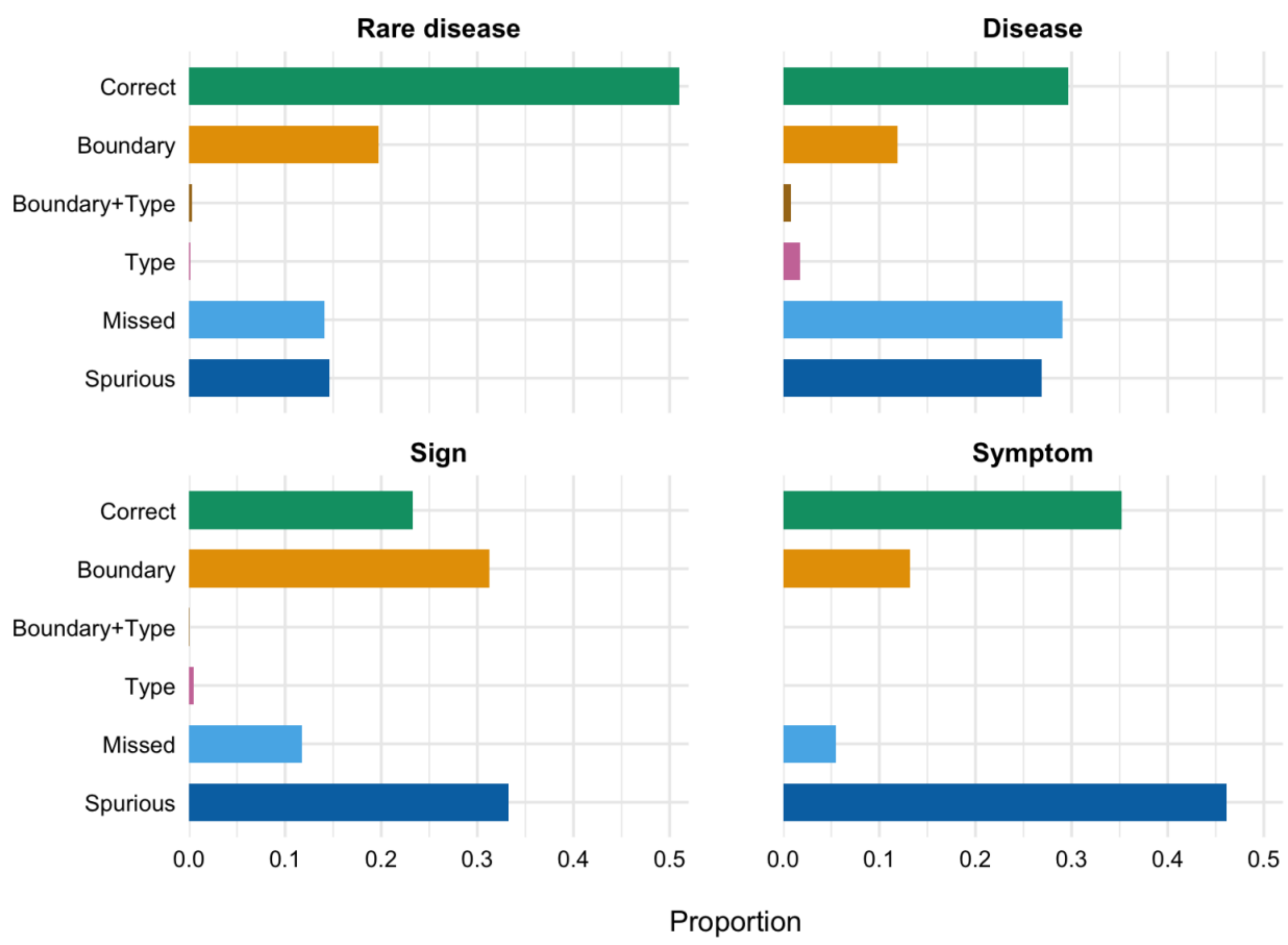

**Figure 2. Error distribution for each entity type.** Each bar represents the proportion of entity predictions falling into one of six mutually exclusive categories on the test set. Results are calculated using Inquiry-KNN methods with the best-performing $k$-shot configuration per entity as determined by F1 score in **Figure 1**.

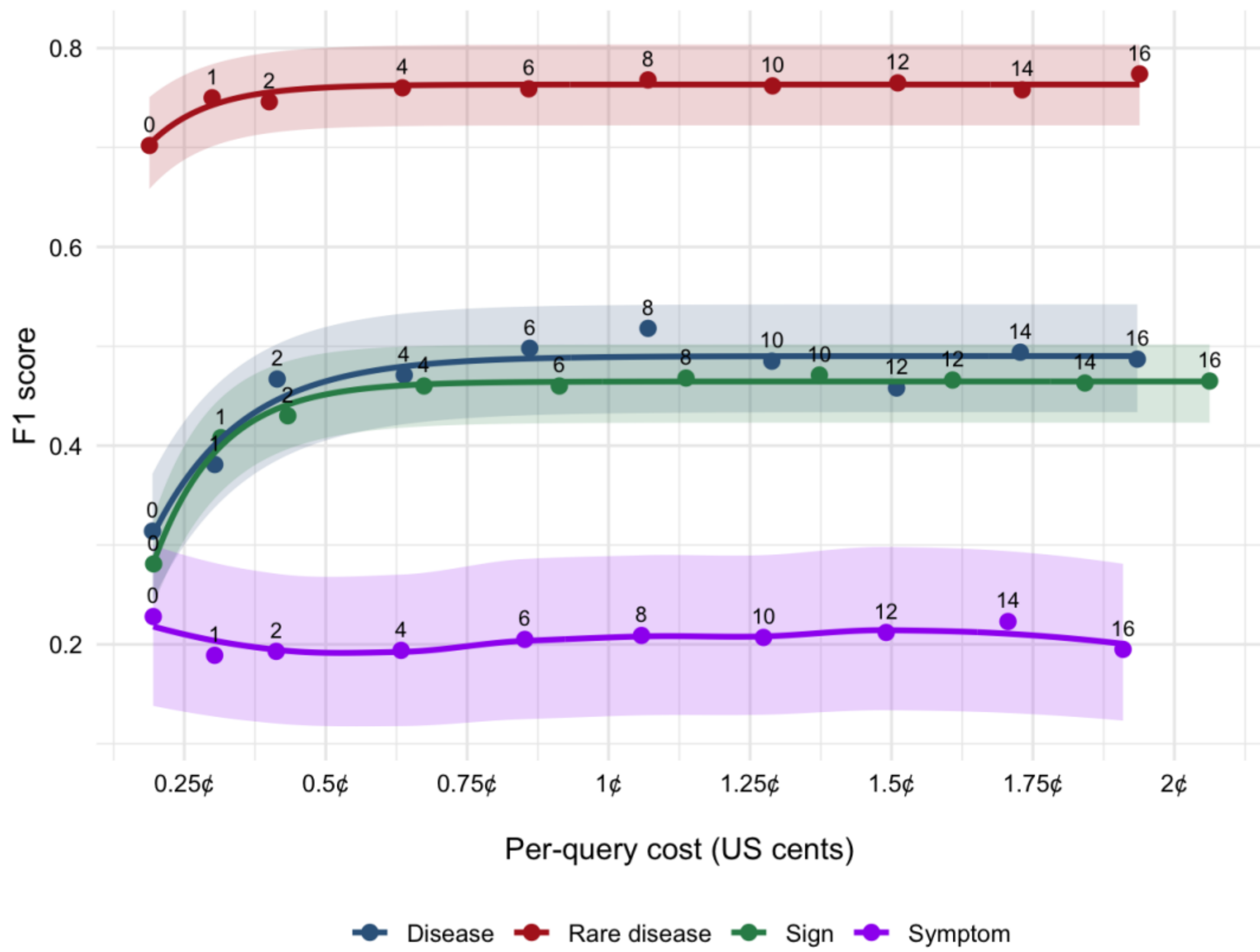

**Figure 3. Cost-performance curves for the four named entities.** Each point corresponds to a $k$-shot prompt evaluated on the test set. Solid lines are entity-specific smoothers: an asymptotic-exponential fit for rare disease, disease and sign, as well as a LOESS smoother for symptom, whose non-monotonic pattern violates the exponential assumption. The color bands show the 95% confidence intervals of F1 scores constructed by bootstrap.

# Supplementary

**Table S1. Paired document-level bootstrap tests for RAG effects.** For each entity type, we report paired differences Δ = (RAG − baseline) in precision, recall, and F1 for two prespecified comparisons: (i) zero-shot + RAG vs zero-shot and (ii) best performing few-shot setting + RAG vs its matched few-shot baseline. Values are bootstrap mean Δ with 95% confidence intervals in brackets. One-sided p-values less than 0.1 are underscored.

| Entity | Comparison | Δ Precision | p | Δ Recall | p | Δ F1 | p |
|---|---|---|---|---|---|---|---|
| Rare Disease | Zero-shot + 2-RAG vs zero-shot | -0.041 (-0.104, 0.026) | 0.896 | -0.121 (-0.168, -0.075) | 1.000 | -0.108 (-0.152, -0.064) | 1.000 |
| | 4-shot + 2-RAG vs 4-shot | 0.030 (-0.012, 0.074) | **0.076** | -0.058 (-0.095, -0.026) | 0.999 | -0.026 (-0.059, 0.002) | 0.965 |
| Disease | Zero-shot + 1-RAG vs zero-shot | -0.060 (-0.141, 0.022) | 0.927 | -0.022 (-0.074, 0.028) | 0.819 | -0.032 (-0.095, 0.028) | 0.850 |
| | 4-shot + 1-RAG vs 4-shot | 0.025 (-0.035, 0.087) | 0.214 | -0.035 (-0.086, 0.018) | 0.916 | -0.011 (-0.059, 0.038) | 0.691 |
| Sign | Zero-shot + 1-RAG vs zero-shot | 0.029 (-0.009, 0.066) | **0.067** | 0.028 (0.006, 0.051) | **0.009** | 0.030 (0.005, 0.055) | **0.010** |
| | 4-shot + 1-RAG vs 4-shot | -0.053 (-0.096, -0.011) | 0.993 | -0.097 (-0.143, -0.051) | 1.000 | -0.076 (-0.116, -0.037) | 1.000 |
| Symptom | Zero-shot + 1-RAG vs zero-shot | -0.032 (-0.062, -0.005) | 0.992 | 0.059 (0.000, 0.148) | 0.134 | -0.041 (-0.083, -0.001) | 0.976 |
| | 4-shot + 1-RAG vs 4-shot | 0.002 (-0.013, 0.019) | 0.448 | 0.038 (0.000, 0.121) | 0.377 | 0.004 (-0.018, 0.032) | 0.403 |

**Table S2. Goodness-of-fit diagnostics for the asymptotic-exponential cost-performance models.** Symptom is excluded because its curve is fit with nonparametric LOESS.

| Entity | n | RMSE | Pseudo $R^2$ |
|---|---|---|---|
| Rare disease | 10 | 0.0058 | 0.9068 |
| Disease | 10 | 0.0171 | 0.9162 |
| Sign | 10 | 0.0052 | 0.9909 |

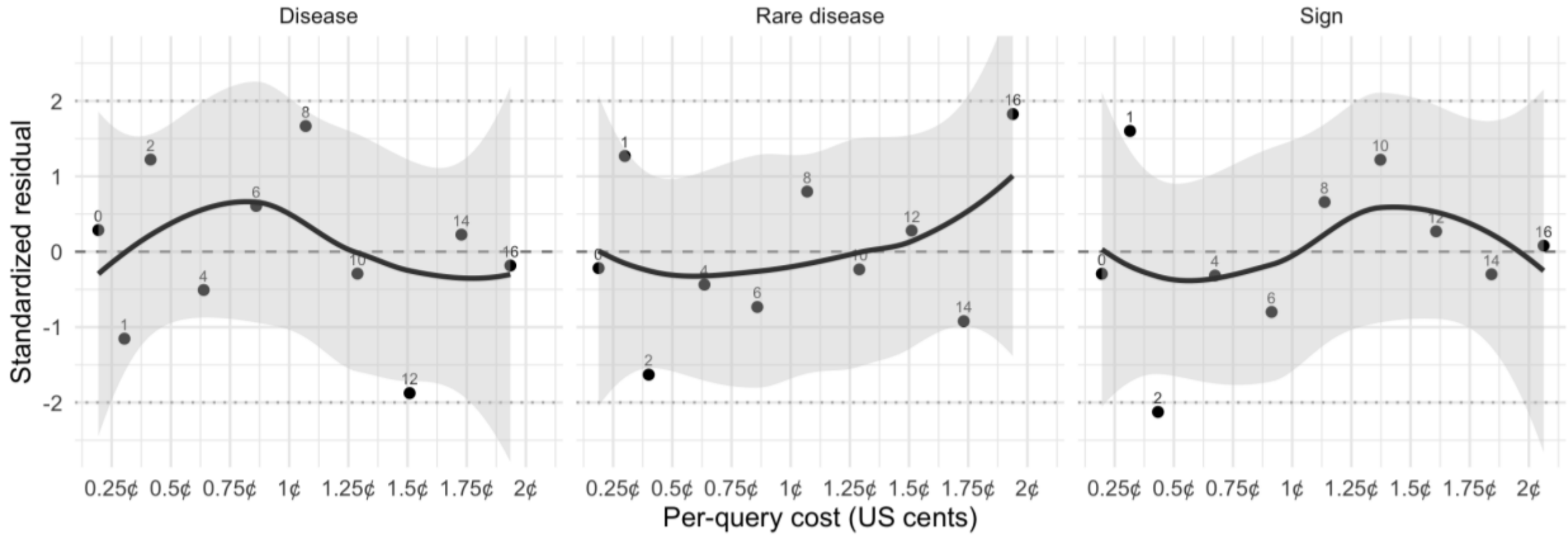

**Figure S1. Residual diagnostics for asymptotic-exponential cost-performance models.** Each panel shows residuals (observed F1−fitted F1) plotted against per-query cost for rare disease, disease, and sign. The dashed horizontal line indicates zero residual. Symptom is excluded because its curve is fitted with nonparametric LOESS.